\documentclass{article}

\PassOptionsToPackage{numbers, compress}{natbib}
\usepackage[final]{timeseries_workshop}
\usepackage{graphicx} 
\usepackage{pythonhighlight}

\usepackage[utf8]{inputenc} % allow utf-8 input
\usepackage[T1]{fontenc}    % use 8-bit T1 fonts
\usepackage{hyperref}       % hyperlinks
\usepackage{url}            % simple URL typesetting
\usepackage{booktabs}       % professional-quality tables
\usepackage{amsfonts}       % blackboard math symbols
\usepackage{nicefrac}       % compact symbols for 1/2, etc.
\usepackage{microtype}      % microtypography
\usepackage{xcolor}         % colors
\usepackage{comment}
\newcommand{\model}[1]{\texttt{#1}}
\newcommand{\TimeCopilot}{\model{{TimeCopilot}}}

\title{TimeCopilot}

\author{%
  Azul Garza\\
  San Francisco, CA, USA \\
  \texttt{azul.garza.r@gmail.com} 
    \And
  Renée Rosillo \\
  San Francisco, CA, USA \\
  \texttt{reneerosillo@protonmail.com } \\
}

\usepackage{listings}
\usepackage{xcolor}

\begin{document}

\maketitle

\begin{abstract}

\begin{comment}
Recent advances in time series forecasting have led to a proliferation of Time Series Foundation Models (TSFMs), each trained on large-scale collections of temporal data and offering different inductive biases and APIs. While these models achieve strong performance, their fragmentation makes it difficult for practitioners to compare, integrate, and deploy them in real-world pipelines. At the same time, Large Language Models (LLMs) have emerged as powerful agents for orchestrating complex workflows across domains.  
\end{comment}

We introduce \TimeCopilot, the first open-source agentic framework for forecasting that combines multiple Time Series Foundation Models (TSFMs) with Large Language Models (LLMs) through a single unified API. \TimeCopilot\ automates the forecasting pipeline: feature analysis, model selection, cross-validation, and forecast generation, while providing natural language explanations and supporting direct queries about the future. The framework is LLM-agnostic, compatible with both commercial and open-source models, and supports ensembles across diverse forecasting families. Results on the large-scale GIFT-Eval benchmark show that \TimeCopilot\ achieves state-of-the-art probabilistic forecasting performance at low cost. Our framework provides a practical foundation for reproducible, explainable, and accessible agentic forecasting systems. 
%with source code at \url{https://github.com/AzulGarza/TimeCopilot} and documentation at \url{https://timecopilot.dev}.
\end{abstract}

\section{Introduction}

The forecasting community has seen a rapid growth of TSFMs \citep{kottapalli2025foundationmodelstimeseries}. These models are typically pre-trained on large collections of heterogeneous time series and then adapted to generate forecasts on unseen datasets \citep{ma2024surveytimeseriespretrainedmodels, oreshkin2021meta, bommasani2022opportunitiesrisksfoundationmodels}. So far, we have found approaches spanning a diverse set of architectures: non-Transformer models such as Tiny Time Mixers \citep{ekambaram2024tinytimemixersttms}; encoder-only designs including Moment \citep{goswami2024momentfamilyopentimeseries} and Moirai \citep{woo2024unifiedtraininguniversaltime}; decoder-only architectures such as Timer-XL \citep{liu2025timerxllongcontexttransformersunified}, Time-MOE \citep{shi2025timemoebillionscaletimeseries}, ToTo \citep{cohen2024tototimeseriesoptimized}, Timer \citep{liu2024timergenerativepretrainedtransformers}, TimesFM \citep{das2023decoder}, and Lag-Llama \citep{rasul2023lag}; approaches that adapt pretrained LLMs to time series forecasting, including Chronos \citep{ansari2024chronoslearninglanguagetime}, AutoTimes \citep{liu2024autotimesautoregressivetimeseries}, LLMTime \citep{gruver2024largelanguagemodelszeroshot}, Time-LLM \citep{jin2023time}, and FPT \citep{lu2021pretrainedtransformersuniversalcomputation}; and models deployed as a web API such as TimeGPT-1 \citep{garza2023timegpt}. Each line of work emphasizes different inductive biases and pretraining strategies, showing strong performance on benchmarks while highlighting trade-offs in scalability, interpretability, and robustness.

While this proliferation of models has expanded the options available to a seasoned forecaster when determining the best approach for a given use case, it has also increased complexity: since a particular laboratory or research group develops each model, each comes with its own API, training pipeline, and evaluation conventions. Beyond these differences, models also vary in their data input requirements (e.g., univariate vs. multivariate, patch vs. non-patch tokenization) and in their learning curve complexity, with some demanding extensive compute and expertise to deploy or use effectively. This fragmentation makes it difficult to compare models fairly, and even harder to integrate them seamlessly into production forecasting systems \citep{Bergmeir2024llms}. Recent large-scale evaluations such as GIFT-Eval \citep{aksu2024giftevalbenchmarkgeneraltime} highlight these challenges by showing in practice how different design choices across foundation models complicate reproducibility and cross-benchmark comparisons.

Meanwhile, the broader AI community has witnessed an explosion in the use of LLMs to orchestrate tasks across domains \cite{brown2020,vaswani2017attention}. In particular, the agentic paradigm, where an LLM serves as a controller that plans, reasons, and executes tasks \citep{liu2024largelanguagemodelbasedagents}, has been increasingly adopted in applications spanning text, code, image, video, and audio generation \citep{sapkota2025aiagentsvsagentic}. Agents allow end-users to specify high-level goals in natural language, while the underlying system orchestrates specialized models and tools to achieve them.

Within time series, however, agentic approaches are still very new. For example, \cite{cai2025timeseriesgymscalablebenchmarktime} introduces a scalable benchmarking framework to evaluate AI agents in machine learning engineering tasks related to time series, such as data handling and code translation, rather than forecasting itself. In parallel, multimodal efforts such as \cite{williams2025contextkeybenchmarkforecasting} highlight another emerging intersection between time series and LLMs: integrating numerical signals with natural language context to produce more reliable forecasts. Together, these efforts underscore both the promise and the nascency of agentic and LLM-augmented approaches in the forecasting domain. Despite these first steps, no prior work has unified TSFMs under an agentic interface.

\begin{figure}[h!]
    \centering
    \begin{minipage}{0.6\textwidth}
        \centering
        \includegraphics[width=\linewidth]{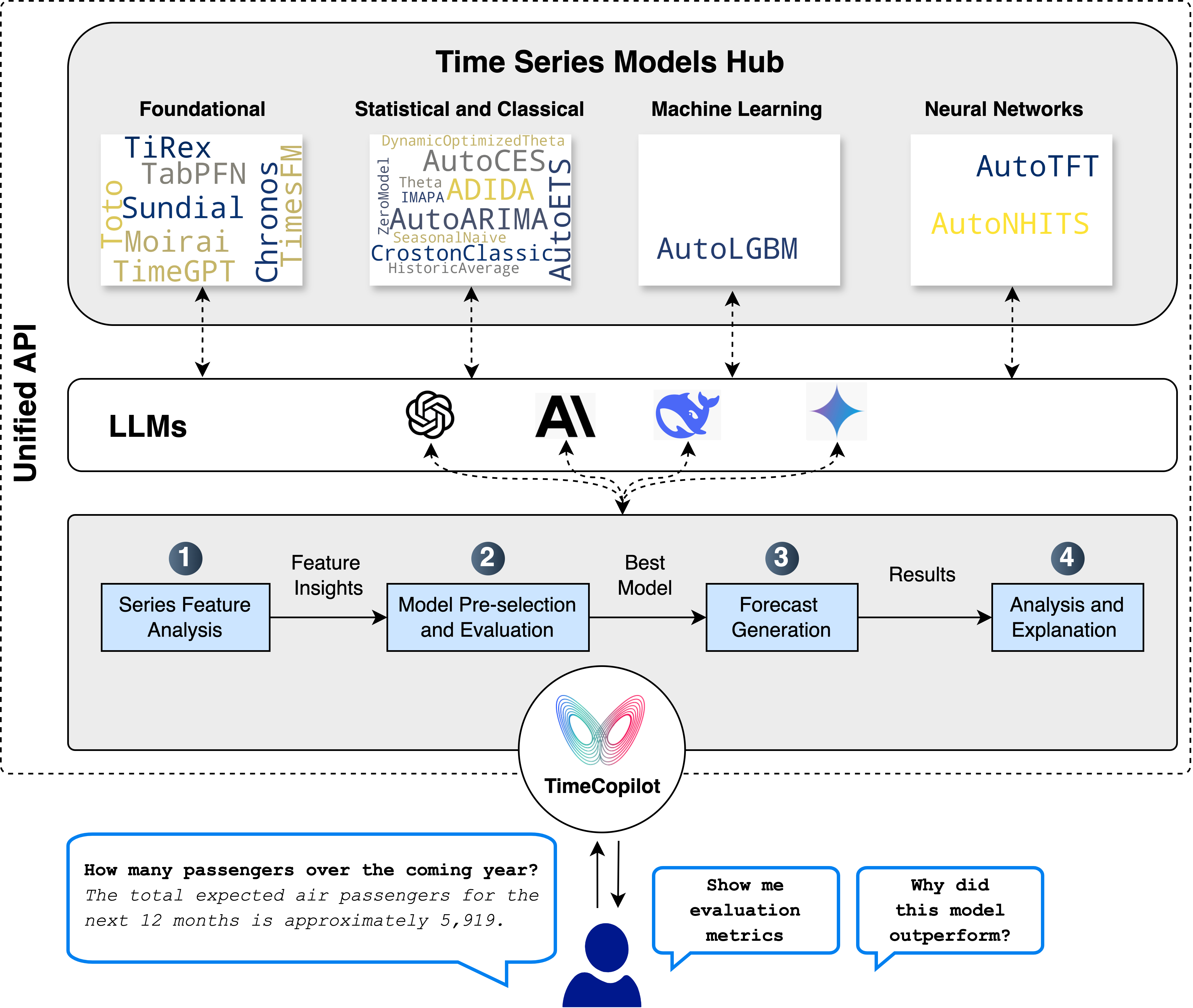}
    \end{minipage}%
    \hfill
    \begin{minipage}{0.30\textwidth}
        \begin{lstlisting}
from timecopilot import TimeCopilot

tc = TimeCopilot(
    llm="openai:gpt-4o"
)

result = tc.forecast(
    df=df,
    query="how many air passengers are expected in the next 12 months?",
)
    
print(result.output.user_query_response)

"The total expected air passengers for the next 12 months is approximately 5,919."
        \end{lstlisting}
    \end{minipage}
    \caption{\footnotesize (Right) Overall \TimeCopilot's architecture. (Left) \TimeCopilot\ Agent API usage.}
    \label{fig:architecture}
\end{figure}

In this paper, we introduce \TimeCopilot\footnote{\TimeCopilot\ is available via \texttt{uv add timecopilot} and documentation at \url{https://timecopilot.dev}.}, the first open-source framework designed specifically for agentic forecasting, see Figure~\ref{fig:architecture}. \TimeCopilot\ is a generative forecasting agent where LLMs act as reasoning engines to orchestrate multiple TSFMs, as well as classical statistical, machine learning, and deep learning methods, through a single unified API. This design eliminates fragmented dependencies across research groups and automates the full forecasting pipeline, from data preparation and model selection to ensembling and evaluation, while providing natural language explanations and supporting direct queries about the future. By bridging the agent paradigm with a unified hub of forecasting models, \TimeCopilot\ provides a practical path toward reproducible, interactive, and accessible agentic forecasting workflows.

\section{Design Principles}

\TimeCopilot\ was designed to automate classical forecasting pipelines \citep{Armstrong01, hyndman2024fpppy}. Traditionally, a forecaster explores a time series, extracts relevant information (features, plots, or diagnostics), formulates hypotheses about which models may perform well, and then uses cross-validation and benchmarking to select the final model \citep{MAKRIDAKIS202054}. This cycle of exploration, hypothesis, and empirical validation has long been considered the cornerstone of forecasting practice.

\TimeCopilot\ leverages LLMs in two ways: (i) to decide how to act at each step of this process, using their capacity to orchestrate multiple tools and reason over diverse signals, and (ii) to help forecasters explain both the decisions made during model selection and the forecasts themselves in natural language, making the results transparent and accessible. This combination of automation and explainability makes \TimeCopilot\ the first agentic interface tailored to time series forecasting.

As Figure~\ref{fig:architecture} shows, the framework is LLM-agnostic: it can interface with commercial APIs (e.g., OpenAI \citep{brown2020}, Anthropic) as well as open-source models such as DeepSeek \citep{deepseekai2025deepseekr1incentivizingreasoningcapability} or LLaMA \citep{touvron2023llamaopenefficientfoundation}, giving users flexibility to choose models according to cost, availability, or deployment constraints.

\subsection{\TimeCopilot\ Agent}

The \textbf{\TimeCopilot\ Agent} orchestrates the forecasting workflow in three structured steps: (i) \textbf{Time Series Feature Analysis:} computes focused diagnostics (trend, seasonality, stationarity) that directly inform model choice \citep{garza2023tsfeatures, hyndman2024tsfeatures} (ii) \textbf{Model Selection and Evaluation:} proposes candidate models starting from simple statistical baselines, documents their assumptions, evaluates them through cross-validation, and escalates to more complex models only when needed. \citep{hyndman2024fpppy} (iii) \textbf{Final Model Selection and Forecasting:} chooses the best model based on metrics, generates forecasts while interpreting patterns, uncertainty, and reliability.  Beyond automating these steps, the agent is designed for explainability: users can query not only the forecasts but also the reasoning behind each decision. This transparency differentiates \TimeCopilot\ from black-box agentic systems and is critical for trust in forecasting applications.

\subsection{\TimeCopilot\ Forecaster}

The \textbf{\TimeCopilot\ Forecaster} executes the models proposed by the agent. It provides access to the \textbf{largest unified hub of TSFMs under a single API}
\footnote{A complete and regularly updated list of models is available at \url{https://timecopilot.dev/model-hub/}},
%\footnote{A complete and regularly updated list of models is available (link omitted for anonymity).}, 
alongside statistical, machine learning, and deep learning methods. This hub removes the need for users to manage conflicting dependencies or learn multiple fragmented interfaces. The current set of supported TSFMs includes Chronos \citep{ansari2024chronoslearninglanguagetime}, Chronos-2 \citep{ansari2025chronos2univariateuniversalforecasting}, FlowState \citep{graf2025flowstatesamplingrateinvariant}, Moirai \citep{woo2024unifiedtraininguniversaltime}, Sundial \citep{liu2025sundialfamilyhighlycapable}, TabPFN \citep{hoo2025tablestimetabpfnv2outperforms}, TiRex \citep{auer2025tirexzeroshotforecastinglong}, TimeGPT \citep{garza2023timegpt}, TimesFM \citep{das2023decoder}, and Toto \citep{cohen2024tototimeseriesoptimized}.  

Alongside foundation models, \TimeCopilot\ integrates a wide suite of \textbf{statistical baselines} such as ADIDA, AutoARIMA, AutoETS, Theta, and SeasonalNaive \citep{garza2022statsforecast}, as well as Prophet \citep{prophet}. For \textbf{machine learning methods}, \TimeCopilot\ includes AutoLGBM \citep{nixtla2023mlforecast}. For \textbf{neural networks}, it incorporates AutoNHITS \citep{challu2023nhits} and AutoTFT \citep{tft} via the NeuralForecast library \citep{olivares2022library_neuralforecast}.  

Finally, \TimeCopilot\ supports \textbf{ensemble techniques}, such as MedianEnsemble, which can combine forecasts from heterogeneous models (statistical, ML, and foundation) to improve robustness \citep{batesgranger1969}. This breadth enables direct comparison and flexible combinations, ensuring forecasters can benchmark and deploy the most suitable approach with minimal effort.

\section{Benchmarks and API Usage Example}

To illustrate the capabilities of \TimeCopilot\, we provide both benchmark results and minimal usage examples. 

\begin{figure}[h!]
  \centering
  \includegraphics[width=1.0\textwidth]{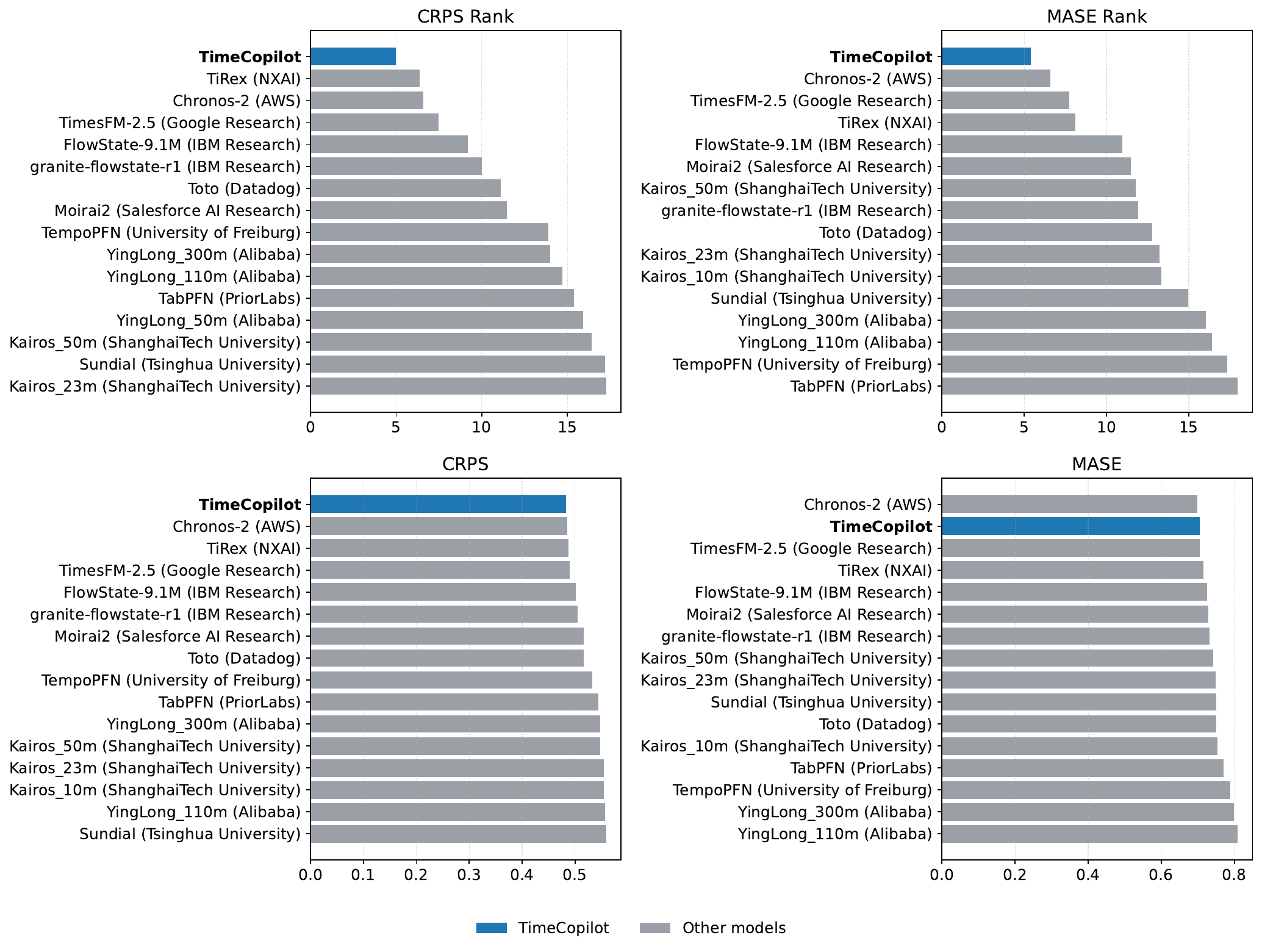}
  \caption{\footnotesize Performance of \TimeCopilot\ and baseline models on the GIFT-Eval benchmark \citep{aksu2024giftevalbenchmarkgeneraltime}. Lower values indicate better forecast performance. The top-16 models are shown, for each metric (CRPS and MASE). The top row shows the mean ranks across datasets, while the bottom row shows the corresponding mean scores.}
  \label{fig:gifteval}
\end{figure}

\subsection{Benchmarks on GIFT-Eval}

Figure~\ref{fig:gifteval} summarizes the mean ranks and scores of foundation models on the GIFT-Eval benchmark \cite{aksu2024giftevalbenchmarkgeneraltime}, which spans 24 datasets, 144k+ time series, and 177M data points across multiple domains and frequencies. The top row reports the average rank across datasets for probabilistic accuracy (CRPS; \cite{Gneiting2014}) and point accuracy (MASE; \cite{HK06}), while the bottom row shows the corresponding mean scores. We include only methods submitted to GIFT-Eval that are publicly reproducible\footnote{Live leaderboard and details available at \url{https://huggingface.co/spaces/Salesforce/GIFT-Eval}}
. An ensemble of foundation models built with \TimeCopilot\ achieved the best overall performance while maintaining reproducibility and low computational cost (approximately \$24 of GPU-distributed inference).  

The \TimeCopilot\ results rely on its MedianEnsemble feature \footnote{\url{https://timecopilot.dev/api/models/ensembles}}.
%\footnote{The ensemble implementation will be released upon acceptance.}.
\citep{PETROPOULOS2020110, batesgranger1969}, used to combine three state-of-the-art foundation models (Chronos-2 \citep{ansari2025chronos2univariateuniversalforecasting}, TimesFM \citep{das2023decoder}, and TiRex \citep{auer2025tirexzeroshotforecastinglong}) with an isotonic regression \citep{Barlow01031972} to ensure monotonic quantiles for probabilistic forecasting. This approach provides robustness against outliers and model-specific biases. %\footnote{The full experiment will be made reproducible upon acceptance.}
\footnote{The full experiment can be reproduced at \url{https://timecopilot.dev/experiments/gift-eval}.}  

These results highlight three aspects of \TimeCopilot: (i) It can orchestrate multiple foundation models through a unified interface. (ii) It provides state-of-the-art results in both probabilistic and point forecasting. (iii) It enables reproducible and cost-effective large-scale experimentation. 

\subsection{API Usage Example}

\TimeCopilot\ provides two main entry points: the \textbf{Agent} (end‑to‑end orchestration with natural language explanations and queries about the future) see Figure~\ref{fig:architecture}, and the \textbf{Forecaster} (direct control of specific models under a unified API), see Appendix~\ref{sec:appendix} for more details. This simple interface allows users to switch seamlessly between agent-driven automation and manual benchmarking of specific models, while keeping the same data structure and evaluation utilities.

\section{Conclusion and Future Work}

We presented \TimeCopilot, the first open-source agentic interface and framework that combines multiple TSFMs with LLMs to make the forecasting workflow explainable, transparent, and accessible to a wider audience. \TimeCopilot\ addresses the challenge of unifying diverse forecasting models under a single API, while leveraging LLMs to automate pipeline decisions and provide natural language explanations for both model choices and forecasts. 

\TimeCopilot's future work encompasses: i) \textbf{Integrating with Model Context Protocol (MCP)} enabling seamless interaction with external tools and services through standardized interfaces, ii) \textbf{Expanding use cases} beyond academic benchmarks to domains such as energy, climate, finance, and supply chain forecasting, and iii) \textbf{Extending forecasting capabilities} to support hierarchical forecasting, multivariate forecasting with multivariate explanations, and advanced evaluation settings such as coherence across multiple aggregation levels.  

\begin{comment}
By bridging LLM agents with state-of-the-art forecasting models, TimeCopilot makes it possible to forecast the future, simply.
\end{comment}

\section*{Acknowledgments}
We would like to express our gratitude to Martín Juárez, Rodrigo Mendoza-Smith, and Salomón Márquez for their helpful contributions and support.

\bibliographystyle{unsrtnat}
\bibliography{references}

\section{Appendix}
\label{sec:appendix}

Code snippet demonstrating the \TimeCopilot\ Forecaster API, which benchmarks and deploys statistical, machine-learning, neural, and foundational models through a single unified interface.

\begin{lstlisting}
from timecopilot import TimeCopilotForecaster
from timecopilot.models.foundation.toto import Toto
from timecopilot.models.ml import AutoLGBM
from timecopilot.models.neural import AutoNHITS
from timecopilot.models.prophet import Prophet
from timecopilot.models.stats import AutoARIMA, SeasonalNaive


df = pd.read_csv(
    "https://timecopilot.s3.amazonaws.com/public/data/air_passengers.csv",
    parse_dates=["ds"],
)

tcf = TimeCopilotForecaster(
    models=[
        AutoARIMA(),
        AutoLGBM(),
        AutoNHITS(),
        Prophet(),
        SeasonalNaive(),
        Toto(),
    ]
)

fcst_df = tcf.forecast(df=df, h=12)
cv_df = tcf.cross_validation(df=df, h=12)
\end{lstlisting}

\end{document}